\documentclass[a4paper,12pt]{article}

\usepackage{fancyvrb}
\usepackage{graphicx}
\usepackage{hyperref}
\usepackage{ragged2e}
\usepackage[section]{placeins}

\begin{document}

\title{Smart machines and the SP theory of intelligence}

\author{J Gerard Wolff\footnote{Dr Gerry Wolff, BA (Cantab), PhD (Wales), CEng, MBCS (CITP); CognitionResearch.org, Menai Bridge, UK; \href{mailto:jgw@cognitionresearch.org}{jgw@cognitionresearch.org}; +44 (0) 1248 712962; +44 (0) 7746 290775; {\em Skype}: gerry.wolff; {\em Web}: \href{http://www.cognitionresearch.org}{www.cognitionresearch.org}.}}

\maketitle

\begin{abstract}

These notes describe how the {\em SP theory of intelligence}, and its embodiment in the {\em SP machine}, may help to realise cognitive computing, as described in the book {\em Smart Machines}. In the SP system, information compression and a concept of {\em multiple alignment} are centre stage. The system is designed to integrate such things as unsupervised learning, pattern recognition, probabilistic reasoning, and more. It may help to overcome the problem of variety in big data, it may serve in pattern recognition and in the unsupervised learning of structure in data, and it may facilitate the management and transmission of big data. There is potential, via information compression, for substantial gains in computational efficiency, especially in the use of energy. The SP system may help to realise data-centric computing, perhaps via a development of Hebb's concept of a `cell assembly', or via the use of light or DNA for the processing of information. It has potential in the management of errors and uncertainty in data, in medical diagnosis, in processing streams of data, and in promoting adaptability in robots.

\end{abstract}


\section{Introduction}

The book {\em Smart Machines: IBM's Watson and the Era of Cognitive Computing} \cite{kelly_hamm_2013} provides a very interesting account of the need for ``a new generation of tools---cognitive technologies---that help us to penetrate complexity and comprehend the world around us so that we can make better decisions and live more successfully and sustainably.'' ({\em ibid.}, Preface).

These notes are about how the {\em SP theory of intelligence} and its embodiment in the {\em SP machine} may help to translate that vision into reality.

I'll first introduce the SP concepts briefly and then try to show how they may facilitate some of the possibilities described in {\em Smart Machines}.

\section{Introduction to the SP theory and SP machine}\label{introduction_to_sp_section}

The SP theory, which has been under development for several years, aims to simplify and integrate concepts across artificial intelligence, mainstream computing and human perception and cognition, with information compression as a unifying theme.

\begin{figure}[!htbp]
\centering
\includegraphics[width=0.5\textwidth]{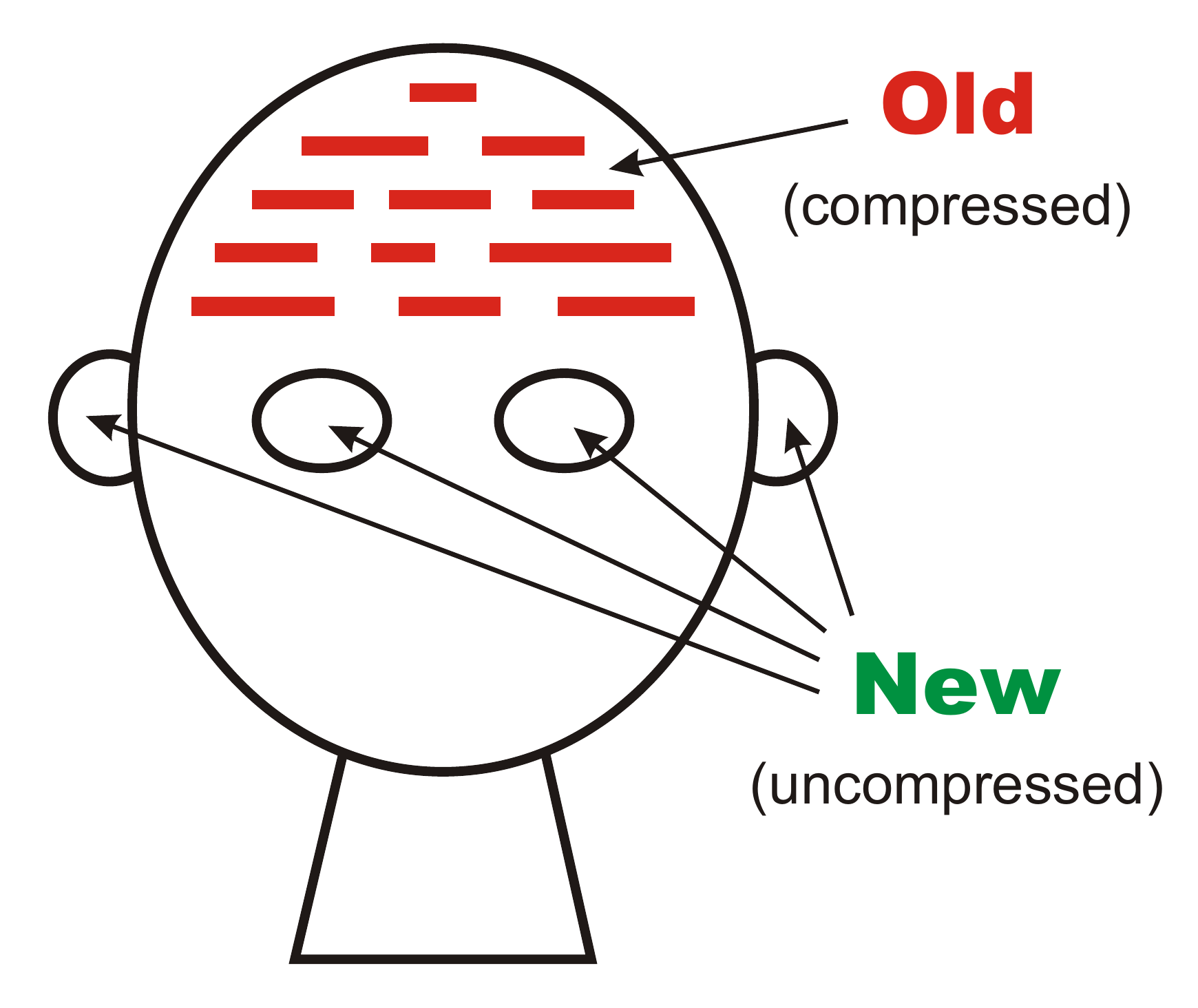}
\caption{Schematic representation of the SP system from an `input' perspective.}
\label{sp_input_perspective_figure}
\end{figure}

The theory is conceived as an abstract brain-like system that, in an `input' perspective, may receive `New' information via its senses, and compress some or all of it to create `Old' information, as illustrated schematically in Figure \ref{sp_input_perspective_figure}. In the theory, information compression is the mechanism both for the learning and organisation of knowledge and for pattern recognition, reasoning, problem-solving, and more.

\begin{figure}[!htbp]
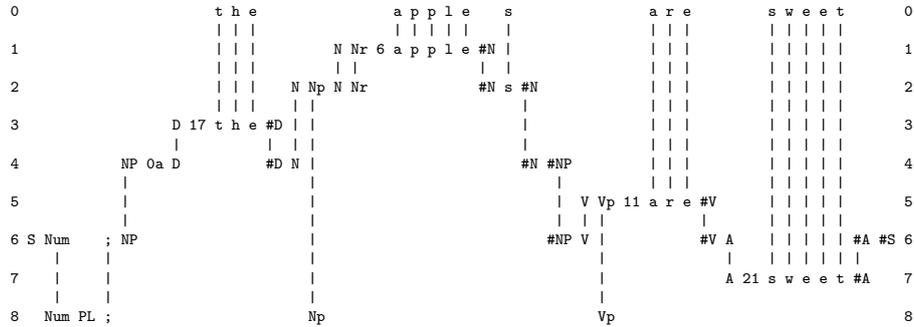

\fontsize{06.00pt}{07.20pt}
\centering
{\bf
\begin{BVerbatim}
0                       t h e                a p p l e    s                a r e         s w e e t       0
                        | | |                | | | | |    |                | | |         | | | | |
1                       | | |         N Nr 6 a p p l e #N |                | | |         | | | | |       1
                        | | |         | |              |  |                | | |         | | | | |
2                       | | |    N Np N Nr             #N s #N             | | |         | | | | |       2
                        | | |    | |                        |              | | |         | | | | |
3                  D 17 t h e #D | |                        |              | | |         | | | | |       3
                   |          |  | |                        |              | | |         | | | | |
4            NP 0a D          #D N |                        #N #NP         | | |         | | | | |       4
             |                     |                            |          | | |         | | | | |
5            |                     |                            |  V Vp 11 a r e #V      | | | | |       5
             |                     |                            |  | |           |       | | | | |
6 S Num    ; NP                    |                           #NP V |           #V A    | | | | | #A #S 6
     |     |                       |                                 |              |    | | | | | |
7    |     |                       |                                 |              A 21 s w e e t #A    7
     |     |                       |                                 |
8   Num PL ;                       Np                                Vp                                  8
\end{BVerbatim}
}
\caption{A multiple alignment created by the SP computer model that achieves the effect of parsing a sentence (``t h e a p p l e s a r e s w e e t'').}
\label{parsing_1_figure}
\end{figure}

In the SP theory, all kinds of knowledge are represented with {\em patterns}: arrays of atomic symbols in one or two dimensions. A key part of the system is a concept of {\em multiple alignment}, like that shown in Figure \ref{parsing_1_figure}.\footnote{The concept of multiple alignment in the SP system (\cite[Section 4]{sp_extended_overview}; \cite[Section 3.4]{wolff_2006}) is borrowed from that concept in bioinformatics, but with important differences.}

\begin{figure}[!htbp]
\centering
\includegraphics[width=0.9\textwidth]{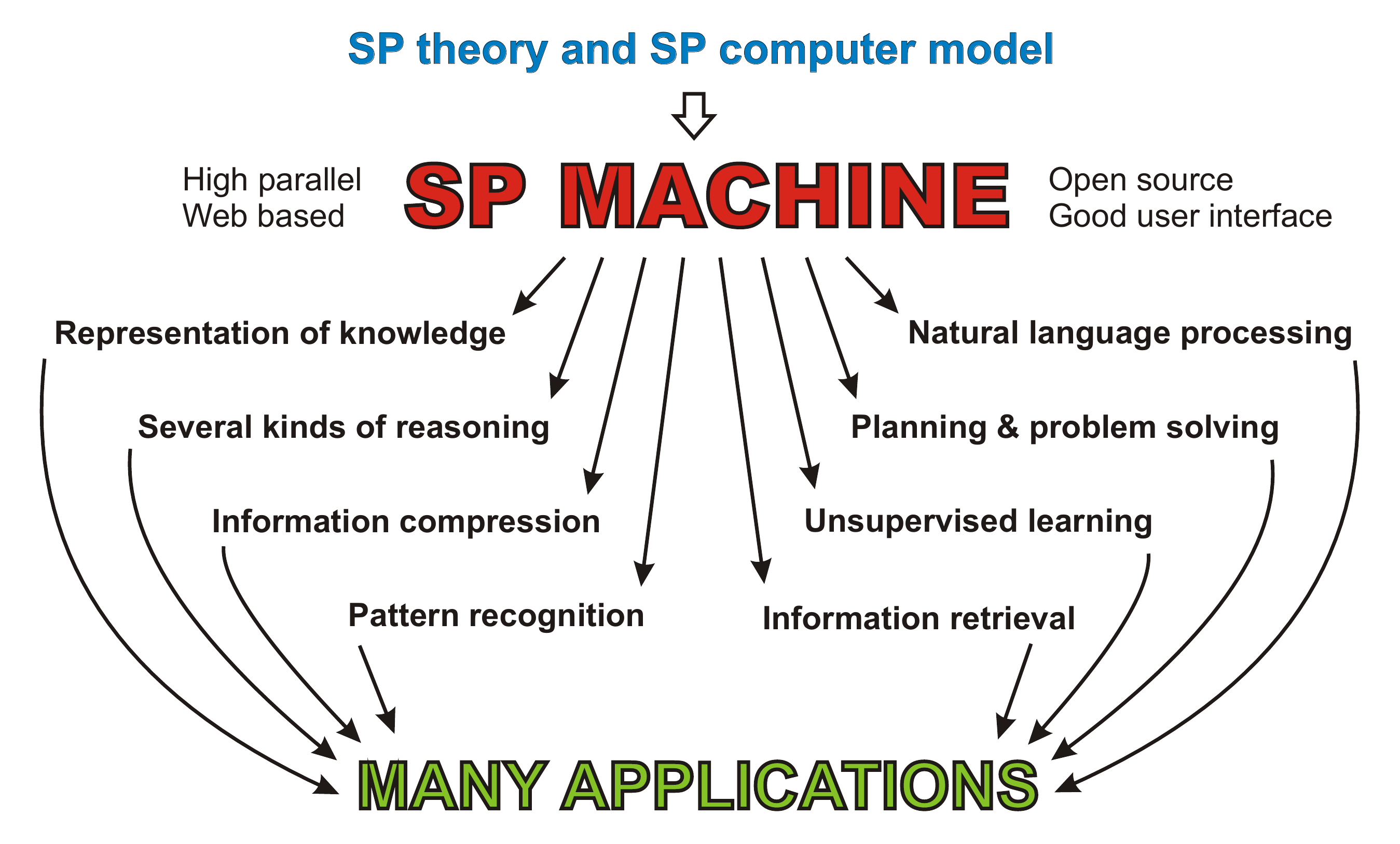}
\caption{Schematic representation of the development and application of the SP machine.}
\label{sp_machine_figure}
\end{figure}

The SP theory is realised in a computer model, SP70, which may be regarded as a first version of the SP machine. It is envisaged that the SP computer model will provide the basis for the development of a high-parallel, open-source version of the SP machine, as shown schematically in Figure \ref{sp_machine_figure}. This will be a means for researchers everywhere to explore what can be done with the system and to create new versions of it.

The SP system provides a model for several aspects of computing and cognition, including unsupervised learning, concepts of computing, aspects of mathematics and logic, the representation of knowledge, natural language processing, pattern recognition, several kinds of reasoning, information storage and retrieval, planning and problem solving, and aspects of neuroscience and of human perception and cognition \cite{sp_extended_overview,wolff_2006}.

\sloppy There is a relatively full account of the SP system in \cite{wolff_2006}, an extended overview in \cite{sp_extended_overview}, an account of its existing and expected benefits and applications in \cite{sp_benefits_apps}, a description of its foundations in \cite{sp_foundations}, and an introduction to the system in \cite{sp_introduction}. There is more information in \href{http://www.cognitionresearch.org/sp.htm}{www.cognitionresearch.org/sp.htm}.

\section{Unsupervised learning}\label{unsupervised_learning_section}

\begin{quote}

``While traditional computers must be programmed by humans to perform specific tasks, cognitive systems will learn from their interactions with data and humans and be able to, in a sense, program themselves to perform new tasks.'' \cite[p.~7]{kelly_hamm_2013}.

\end{quote}

The SP programme derives in part from earlier research developing computer models of language learning.\footnote{\raggedright See \href{http://bit.ly/JCd6jm}{bit.ly/JCd6jm}.}

At a fundamental level---compression of information via the matching and unification of patterns---the SP computer model works in the same way as the earlier models of language learning. Like the earlier models, the SP computer model is able to discover generative grammars from data, including segmental structure, classes of structure, and abstract patterns.

But in developing the SP system, a radical reorganisation has been needed to meet the goal of simplifying and integrating concepts across artificial intelligence, mainstream computing, and human perception and cognition. Unlike the earlier models, multiple alignment is central in the workings of the SP computer model, including unsupervised learning. A bonus of the new structure is potential for the unsupervised learning of class hierarchies, part-whole hierarchies, and discontinuous dependencies in data.

The emphasis on unsupervised learning in the SP programme of research is because of evidence that human learning, including the learning of natural languages, does not depend on the provision of a `teacher', or anything equivalent---although such assistance may be helpful.

A strength of the SP system is that it can discover significant entities, classes of entity, and other structures in data, not merely statistical associations between pre-established structures. An important idea in the SP programme is the `DONSVIC' principle \cite[Section 5.2]{sp_extended_overview}: the conjecture, supported by evidence, that information compression is the key to the discovery of `natural' structures, meaning the kinds of things that people naturally recognise, such as words, objects, and classes of objects.

Learning in the SP system has much of the flavour of:

\begin{quote}

``[building and maintaining] a giant jigsaw puzzle on the fly, making connections between newly acquired data and older information.'' \cite[p.~55]{kelly_hamm_2013}.

\end{quote}

\noindent and of:

\begin{quote}

``... finding many needles in a field full of haystacks, combining those needles, then searching the smaller but still substantial collection for the needles made of gold---the truly valuable slivers of insight.'' \cite[p.~56]{kelly_hamm_2013}.

\end{quote}

\section{Big data}\label{big_data_section}

\begin{quote}

``The emergence of social networking, sensor networks, and huge storehouses of business, scientific, and government records creates an abundance of information tech-industry insiders call `big data'. ... This digital universe is growing at about 60 percent each year.'' \cite[p.~5]{kelly_hamm_2013}.

\end{quote}

\begin{quote}

``We need a new generation of data storage, management, and analytics tools that will improve our ability to gather, meld, and make sense of huge amounts of data and, in some cases, to perform a complex ballet of tasks in real time.'' \cite[p.~45]{kelly_hamm_2013}.

\end{quote}

The SP system may assist in the management and analysis of big data in four main ways:

\begin{itemize}

\item {\em Helping to overcome the problem of variety in big data}.

\begin{quote}

``The manipulation and integration of heterogeneous data from different sources into a meaningful common representation is a major challenge.'' \cite[p.~76]{national_research_council_2013}.\footnote{\raggedright See also the problem of variety described in \cite{laney_2001} and \cite[p.~45]{kelly_hamm_2013}}

\end{quote}

 The versatility of SP patterns, within the multiple alignment framework, in the representation and processing of diverse kinds of knowledge (\cite[Sections 5 to 13]{sp_extended_overview}; \cite[Chapters 5 to 10]{wolff_2006}) makes them a contender for the role of ``meaningful common representation'' or universal format for knowledge. There is potential for the translation of heterogeneous data into that universal format, for harmonisation in the representation of different kinds of knowledge, and for automatic structuring of knowledge via processes for unsupervised learning (Section \ref{unsupervised_learning_section}).

\item {\em Interpretation}. In the interpretation of data, the SP system has capabilities that include such things as natural language processing, several kinds of reasoning, planning, and problem solving. It has several strengths in pattern recognition, where recognition may be achieved: at multiple levels of abstraction; with ``family resemblance'' or polythetic categories; in the face of errors of omission, commission or substitution in data; with the calculation of a probability for any given identification, classification or associated inference; with sensitivity to context in recognition; and with the seamless integration of pattern recognition with other aspects of intelligence---reasoning, learning, problem solving, and so on (\cite[Section 9]{sp_extended_overview}; \cite[Chapter 6]{wolff_2006}).

\item {\em Unsupervised learning}. As outlined in Section \ref{unsupervised_learning_section}, unsupervised learning is a key feature of the SP system. With big data, there is clear potential for the previously-mentioned DONSVIC principle: {\em the discovery of natural structures via information compression} \cite[Section 5.2]{sp_extended_overview}.

\item {\em Information compression}. Information compression is central in how the SP system works. Potential benefits with big data include:

\begin{itemize}

\item Reducing the volume of data and thus facilitating its storage and management. In that connection, there are reasons to believe that the system may achieve higher levels of compression than conventional systems for information compression \cite[Section 6.7]{sp_benefits_apps}.

\item By sending only `code' and not `grammar', there is potential for very substantial economies in the transmission of big data \cite[Section 6.7.1]{sp_benefits_apps}. There is more about this in Section \ref{economies_in_transmission_section}.

\item It appears that information compression can mean substantial gains in computational efficiency, especially in the use of energy (Section \ref{efficiency_ic_section}).

\item In the SP system, information compression provides a key to the management of errors and uncertainty in data (Section \ref{errors_uncertainties_section}).

\item Knowledge structures created by the system, including class-inclusion hierarchies, part-whole hierarchies, and their integration \cite[Section 9.1]{sp_extended_overview}, may facilitate retrieval of information by serving, in effect, as hierarchical indices.

\end{itemize}

\end{itemize}

How the SP system may assist in the management and analysis of big data will be described more fully in a new article \cite{sp_big_data}.

\section{Computational efficiency via information compression}\label{efficiency_ic_section}

\begin{quote}

``... we're reaching the limits of our ability to make [gains in the capabilities of CPUs] at a time when we need even more computing power to deal with complexity and big data. And that's putting unbearable demands on today's computing technologies---mainly because today's computers require so much energy to perform their work.'' \cite[p.~9]{kelly_hamm_2013}.

\end{quote}

\begin{quote}

``The human brain is a marvel. A mere 20 watts of energy are required to power the 22 billion neurons in a brain that's roughly the size of a grapefruit. To field a conventional computer with comparable cognitive capacity would require gigawatts of electricity and a machine the size of a football field. So, clearly, something has to change fundamentally in computing for sensing machines to help us make use of the millions of hours of video, billions of photographs, and countless sensory signals that surround us. ... Unless we can make computers many orders of magnitude more energy efficient, we're not going to be able to use them extensively as our intelligent assistants.'' \cite[p.~75, p.~88]{kelly_hamm_2013}.

\end{quote}

In the quest for greater efficiency in processing, especially energy efficiency, the SP system may make what is potentially a very substantial contribution via information compression (this section) and via data-centric computing (Section \ref{data-centric_computing_section}).

Since information processing in the SP system means compression of information via the matching and unification of patterns, {\em anything that increases the efficiency of searching for good full and partial matches between patterns will also increase the efficiency of information processing}.

It appears that information compression can itself be a means of increasing the efficiency of searching, as described in the next two subsections. Information compression may also yield economies in the transmission of information (Section \ref{economies_in_transmission_section}, below).

\subsection{Reducing the sizes of data to be searched and of search terms}\label{reducing_data_sizes_section}

As described in \cite[Section 6.7.2]{sp_benefits_apps}, if we wish to search a body of information, $I$, for instances of a pattern like ``Treaty on the Functioning of the European Union'' the efficiency of searching may be increased:

\begin{itemize}

\item By reducing the size of $I$ so that there is less to be searched. The size of $I$ may be reduced by replacing all but one of the instances of ``Treaty on the Functioning of the European Union'' with a relatively short code like ``TFEU'', and likewise other recurrent patterns. More generally, the size of $I$ may be reduced via the compression processes in the SP system.

\item By searching with a short code like ``TFEU'' instead of a relatively large pattern like ``Treaty on the Functioning of the European Union''. Other things being equal, a smaller search pattern means more efficient searching.

\end{itemize}

With regard to the second point, there is potential to cut out some searching altogether by ``hard wiring'' the connection between each instance of a code (``TFEU'' in this example) and the thing that it represents (``Treaty on the Functioning of the European Union''). In {\em SP-neural} (Section \ref{sp_neural_section}), there are connections of that kind between ``pattern assemblies'', as shown schematically in Figure \ref{class_part_figure}.

\subsection{Concentrating search where good results are most likely to be found}\label{concentrating_search_section}

If we want to find some strawberry jam, our search is more likely to be successful in a supermarket than it would be in an antiques shop or a showroom for second-hand cars. This may seem too simple and obvious to deserve comment but it illustrates the extraordinary knowledge that most people have of an informal `statistics' of the world that we inhabit, and how that knowledge may help us to minimise effort.\footnote{\raggedright See also G.~K.~Zipf's {\em Human Behaviour and the Principle of Least Effort} \cite{zipf_1949}.}

Where does that statistical knowledge come from? In the SP theory, it flows directly from the central role of information compression in our perceptions, learning and thinking, and from the intimate relationship between information compression and concepts of prediction and probability \cite{li_vitanyi_2009}.

Although the SP computer model calculates probabilities for some purposes (see Section \ref{errors_uncertainties_section}), it actually uses levels of information compression as a guide to search. Those levels are used, with heuristic search methods (including escape from `local peaks'), to ensure that searching is concentrated in areas where it is most likely to be fruitful \cite[Sections 3.9, 3.10, and 9.2]{wolff_2006}. This not only speeds up processing but yields Big-O values for computational complexity that are within acceptable limits \cite[Sections 3.10.6, 9.3.1, and A.4]{wolff_2006}.

\subsection{Economies in the transmission of data}\label{economies_in_transmission_section}

\begin{quote}

``To control costs, designers of the [DOME] computing system have to figure out how to minimize the amount of energy used for processing data. At the same time, since so much of the energy in computing is required to move data around, they have to discover ways to move the data as little as possible.'' \cite[p.~65]{kelly_hamm_2013}.

\end{quote}

Although this quote may refer in part to movements of data such as those between the CPU and the memory of a computer, the discussion here is about transmission of data over longer distances such as, for example, via the internet.

As was mentioned in Section \ref{big_data_section}, there is potential with the SP system for very substantial economies in the transmission of data \cite[Section 6.7.1]{sp_benefits_apps}. Any body of data, $I$, may be compressed by encoding it in terms a `grammar' ($G$), provided that $G$ contains the kinds of structures that are found in $I$ (Section \ref{errors_uncertainties_section}). Then $I$ may be sent from A to B by sending only the `encoding' ($E$). Provided that B has a copy of $G$, $I$ may be recreated with complete fidelity by means of the SP system (\cite[Section 4.5]{sp_extended_overview}; \cite[Section 3.8]{wolff_2006}). Since $E$ would normally be very much smaller than the $I$ from which it was derived, it seems likely that there would be a net gain in efficiency, allowing for the computational costs of encoding and decoding.

Since a copy of $G$ must be transmitted to B, any savings will be relatively small if it is used only for the decoding of a single instance of $E$. But significant savings are likely if, as would normally be the case, one copy of $G$ may be used for the decoding of many different instances of $E$, representing many different bodies of information.

\subsection{Potential gains in computational efficiency}

No attempt has yet been made to quantify potential gains in computational efficiency from the compression of information, as described in Sections \ref{reducing_data_sizes_section}, \ref{concentrating_search_section}, and \ref{economies_in_transmission_section}, but they could be very substantial:

\begin{itemize}

\item Since information compression is fundamental in the workings of the SP system, there is potential for corresponding savings in all parts and levels in the system.

\item The entire structure of knowledge that the system creates for itself is intrinsically statistical, with potential on many fronts for corresponding savings in computational costs and associated demands for energy.

\end{itemize}

It is anticipated that the proposed high-parallel version of the SP machine (Section \ref{introduction_to_sp_section}) will provide a means of exploring these aspects of the system.

\section{Data-centric computing}\label{data-centric_computing_section}

\begin{quote}

``What's needed is a new architecture for computing, one that takes more inspiration from the human brain. Data processing should be distributed throughout the computing system rather than concentrated in a CPU. The processing and the memory should be closely integrated to reduce the shuttling of data and instructions back and forth.'' \cite[p.~9]{kelly_hamm_2013}.

\end{quote}

\begin{quote}

``Scientists at IBM Research believe that to make computing sustainable in the era of big data, we will need a different kind of machine---the data-centric computer. ... Machines will perform computations faster, make sense of large amounts of data, and be more energy efficient.'' \cite[p.~88]{kelly_hamm_2013}.

\end{quote}

The SP concepts may help to integrate processing and memory, as described in the next two subsections.

\subsection{SP-neural}\label{sp_neural_section}

Although the main emphasis in the SP programme has been on developing an abstract framework for the representation and processing of knowledge, the theory includes proposals---called {\em SP-neural}---for how those abstract concepts may be realised with neurons \cite[Chapter 11]{wolff_2006}.

\begin{figure}[!hbt]
\centering
\includegraphics[width=0.8\textwidth]{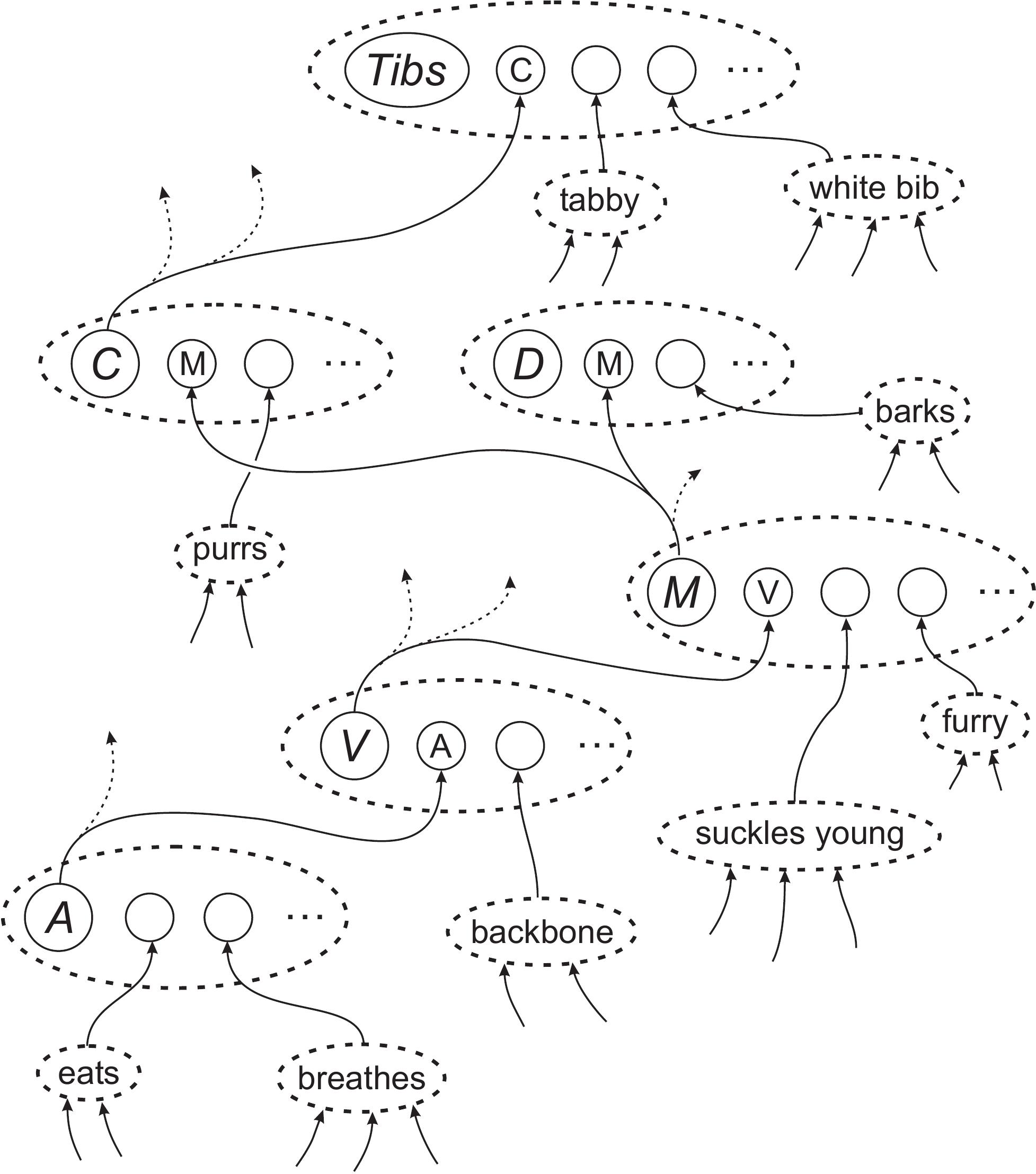}
\caption{An example showing schematically how SP-neural may represent class-inclusion relations, part-whole relations, and their integration. {\em Key}: `C' = cat, `D' = dog, `M' = mammal, `V' = vertebrate, `A' = animal, `...' = further structure that would be shown in a more comprehensive example. Pattern assemblies are surrounded by broken lines and each neuron is represented by an unbroken circle or ellipse. Lines with arrows show connections between pattern assemblies and the flow of sensory signals in the process of recognising something (there may also be connections in the opposite direction to support the production of patterns). Connections between neurons within each pattern assembly are not marked.}
\label{class_part_figure}
\end{figure}

Figure \ref{class_part_figure} shows in outline how an SP-style conceptual structure would appear in SP-neural. It is envisaged that SP patterns would be realised with {\em pattern assemblies}---groupings of neurons like those shown in the figure within broken-line envelopes.

The whole scheme is quite different from `artificial neural networks' as they are commonly conceived in computer science.\footnote{\raggedright See, for example, ``Artificial neural network'', Wikipedia, \href{http://en.wikipedia.org/wiki/Artificial\_neural\_network}{en.wikipedia.org/wiki/Artificial\_neural\_network}, retrieved 2013-12-23.} It may be seen as a development of Donald Hebb's \cite{hebb_1949} concept of a `cell assembly', with more precision about how structures may be shared, and other differences.\footnote{\raggedright In particular, unsupervised learning in the SP system (\cite[Section 5]{sp_extended_overview}; \cite[Chapter 9]{wolff_2006}) is radically different from the ``Hebbian'' concept of learning (see, for example, ``Hebbian theory'', Wikipedia, \href{en.wikipedia.org/wiki/Hebbian\_learning}{http://en.wikipedia.org/wiki/Hebbian\_learning}, retrieved 2013-12-23), described by Hebb \cite{hebb_1949} and adopted as the mechanism for learning in most artificial neural networks. By contrast with Hebbian learning, the SP system, like a person, may learn from a single exposure to some situation or event. And, by contrast with Hebbian learning, it takes time to learn a language in the SP system because of the complexity of the search space, not because of any kind of gradual strengthening or ``weighting'' of links between neurons \cite[Section 11.4.4]{wolff_2006}.}

In SP-neural, the structure of knowledge is reflected directly in groupings of neurons and their interconnections, as shown in the figure. Although the details have not been worked out, it is envisaged that such things as pattern recognition would be achieved via the transmission of impulses between pattern assemblies, and via the transmission of impulses between neurons within each pattern assembly. In keeping with what is known about the workings of brains and nervous systems, it is likely that there would be important roles for both excitatory and inhibitory signals.

In short, neurons in SP-neural serve for both the representation and processing of knowledge, with close integration of the two---in accordance with the concept of data-centric computing.

\subsection{Computing with light or chemicals}

The SP concepts appear to lend themselves to computing with light or chemicals, perhaps by-passing such things as transistors or logic gates that have been prominent in the development of electronic computers \cite[Section 6.10.6]{sp_benefits_apps}.\footnote{``The most promising means of moving data faster is by harnessing photonics, the generation, transmission, and processing of light waves.'' \cite[p.~93]{kelly_hamm_2013}.}

At the heart of the SP system is a process of finding good full and partial matches between patterns. This may be done with light, with the potential advantage that light beams may cross each other without interference. Another potential advantage is that, with collimated light, there may be relatively small losses over distance---although distances should probably be minimised to save on transmission times and to minimise the sizes of computing devices. There appears to be potential to create an optical or optical/electronic version of SP-neural.

Finding good full and partial matches between patterns may also, potentially, be done with chemicals such as DNA,\footnote{See, for example, ``DNA computing'', Wikipedia, \href{http://bit.ly/1gfEP4p}{bit.ly/1gfEP4p}, retrieved 2013-12-30.} with potential for high levels of parallelism, and with the attraction that DNA can be a means of storing information in a very compact form, and for very long periods \cite{goldman_etal_2013}.

With both optical and chemical versions of the SP system, there seems to be potential for achieving data-centric integration of knowledge and processing.

\section{Errors and uncertainties in data}\label{errors_uncertainties_section}

\begin{quote}

``Organizations face huge challenges as they attempt to get their arms around the complex interactions between natural and human-made systems. The enemy is uncertainty. In the past, since computing systems didn't handle uncertainty well, the tendency was to pretend that it didn't exist. Today, it is clear that that approach won't work anymore. So rather than trying to eliminate uncertainty, people have to embrace it.'' \cite[pp.~50--51]{kelly_hamm_2013}.

\end{quote}

The SP system has potential in the management of errors and uncertainties in data. In summary:

\begin{itemize}

\item Owing to the previously-mentioned close connection between information compression and concepts of prediction and probability, the whole system is inherently probabilistic. Every SP pattern has an associated frequency of occurrence. From that information, a probability may be derived for each multiple alignment and for each of what are normally several inferences that may be derived from each multiple alignment (\cite[Section 4.4]{sp_extended_overview}; \cite[Section 3.7 and Chapter 7]{wolff_2006}).\footnote{Although the SP system is fundamentally probabilistic, it can, if required, be constrained to yield all-or-nothing results, much as in a conventional computer \cite[Chapter 10]{wolff_2006}.}

\item Unsupervised learning via information compression provides a neat solution to the problem of ``dirty data'' in the learning of a natural language (\cite[Section 5.3]{sp_extended_overview}; \cite{wolff_1988}): how it is that we can develop a keen sense of what does or does not belong in our native language or languages, despite the fact that most of the speech that children hear contains the kinds of haphazard errors that people make in talking, and in the face of evidence that language learning may be achieved without the benefit of error correction by a teacher, or anything equivalent.

    In brief, the product of learning comprises a `grammar' ($G$) and an `encoding' ($E$) of the original data in terms of the grammar. The two together achieve lossless compression of the original data, including all the errors. But $G$, which may be seen to capture the essence of the target language, excludes everything which is rare---which is mostly those haphazard errors that people make in speaking (these are rare individually, although collectively they are quite common). Anything that is a little more frequently-occurring than rare may acquire the status of linguistic irregularity---such as `bought' (not `buyed') as the past tense of `buy'---or it may be seen as a dialect form.

    These principles apply to any kind of data, not just linguistic data.

\item In operations such as parsing natural language or pattern recognition, the SP system is robust in the face of errors of omission, of commission, or of substitution (\cite[Section 4.2.2]{sp_extended_overview}; \cite[Section 6.2.1]{wolff_2006}). In the same way that we can recognise things visually despite disturbances such as falling leaves or snow, or our own blinking (and likewise for other senses), the SP system can cope quite well with inputs that are not totally correct.

\end{itemize}

Although there are advantages in using probabilities, the flipside---for both people and machines---is that mistakes can be made. We may bet on ``Dancing Queen'' but find that ``Kiss me Kate'' is the winner. In the same way that people can be fooled by a frequently-repeated lie, any probabilistic machine will be vulnerable to systematic distortions in data.

\section{Other possibilities}

Outline here are some other aspects of the SP system, and how they may help to realise the possibilities described in {\em Smart Machines}.

\subsection{Medical diagnosis}

\begin{quote}

``One of the next challenges for Watson is to help doctors diagnose diseases and assess the best treatments for individual patients.'' \cite[p.~2]{kelly_hamm_2013}.

\end{quote}

How the SP system may assist with medical diagnosis is described in \cite{wolff_medical_diagnosis}.

\subsection{Processing streams of data}

\begin{quote}

``Increasingly, vitally important insights can be gained from analyzing information that's on the move. ... Rather than placing the data in a database first, the computer analyzes it as it comes in from a variety of sources, continually refining its understanding of the data as conditions change. This is the way humans process information.'' \cite[pp.~49--50]{kelly_hamm_2013}.

\end{quote}

In its overall organisation, the SP system is designed to process streams of ``New'' information, as shown schematically in Figure \ref{sp_input_perspective_figure}, and very much in the spirit of the quotation above.

As indicated in Section \ref{big_data_section}, the SP system may also provide a handle on the problem of variety in data, mentioned in the quotation.

\subsection{The need for radical innovations}

\begin{quote}

``Soon incremental innovation will no longer be sufficient. ... We need more radical innovations.'' \cite[pp.~17--18]{kelly_hamm_2013}.

\end{quote}

The idea that all kinds of computing and cognition may be understood as information compression via multiple alignment is indeed a radical innovation. But it is not merely an eccentric blind alley. Detailed research has shown its potential in several areas with a bearing on the creation of smart machines.

\subsection{Augmenting our senses}\label{augmenting_senses_section}

\begin{quote}

``... most images and audio on the web are searchable only via metadata---words that are manually typed into forms by humans. So its vital to develop systems that can recognize images and sounds more like humans do.'' \cite[p.~69]{kelly_hamm_2013}.

\end{quote}

As was mentioned in Section \ref{big_data_section}, the SP system has several strengths in pattern recognition (see also \cite[Section 9]{sp_benefits_apps}; \cite[Chapter 6]{wolff_2006}). How the system may be applied to vision is described in \cite{sp_vision}. Because of the generality and versatility of the system \cite[Section 4]{sp_benefits_apps}, it can probably be applied in the recognition of sounds and other kinds of sense data.

\subsection{Achieving human-like flexibility and adaptability in robots}

\begin{quote}

``... as of now, robots remain firmly in the von Neumann computing paradigm. They must be programmed in advance by people to deal with nearly every situation they encounter.'' \cite[p.~72]{kelly_hamm_2013}.

\end{quote}

How, via the SP system, robots may achieve human-like flexibility and adaptability is outlined in \cite[Section 6.3]{sp_benefits_apps}. With robots, especially autonomous robots, the SP system has at least two other attractions: versatility in different aspects of intelligence; and potential for computational efficiency, especially energy efficiency (Sections \ref{efficiency_ic_section} and \ref{data-centric_computing_section}).

\subsection{The need for integration}

\begin{quote}

``Today, as scientists labor to create machine technologies to augment our senses, there's a strong tendency to view each sensory field in isolation as specialists focus only on a single sensory capability. Experts in each sense don't read journals devoted to the others senses, and they don't attend one another's conferences. Even within IBM, our specialists in different sensing technologies don't interact much. Yet if machines are to help humans understand the world, they have to make sense of it and communicate about it in a way that's familiar and comprehensible to humans. This integration of data from various sensing technologies is beginning to happen in multimedia and visual analytics, where vision and sound are correlated. But that's just the start of what will be required in the next era of computing.'' \cite[p.~74]{kelly_hamm_2013}.

\end{quote}

This quotation touches on an issue which is right at the heart of the SP programme of research: the need for simplification and  integration of concepts across different fields. The SP theory aims to apply this principle across artificial intelligence, mainstream computing, and human perception and cognition. Broadening the context in this way improves the chances of creating a theory that, in accordance with Occam's Razor, combines conceptual {\em simplicity} with descriptive or explanatory {\em power} \cite[Section 2]{sp_benefits_apps}.

\section{Conclusion}

The SP system may help to realise cognitive computing, as described in {\em Smart Machines}. It is designed to simplify and integrate concepts across artificial intelligence, mainstream computing, and human perception and cognition. It has potential for unsupervised learning of structure in data, and it may help in the management and analysis of big data. Information compression can mean substantial gains in computational efficiency, especially in the use of energy. The system may help to realise data-centric computing. And it may facilitate the management of errors and uncertainty in data.

The creation of a high-parallel, open-source version of the SP machine, as outlined in Section \ref{introduction_to_sp_section}, would be a means for researchers everywhere to explore what can be done with the system and to create new versions of it.

\bibliographystyle{plain}

\end{document}